%% file: paper.tex
\documentclass[10pt,twocolumn,letterpaper]{article}

\usepackage{cvpr}
\usepackage{times}
\usepackage{epsfig}
\usepackage{graphicx}
\usepackage{amsmath}
\usepackage{amssymb}
\input{symbols.tex}

\usepackage[pagebackref=true,breaklinks=true,letterpaper=true,colorlinks,bookmarks=false]{hyperref}

\cvprfinalcopy 

\ifcvprfinal\fi
\begin{document}

\title{FutureMapping: The Computational Structure of Spatial AI Systems}

\author{Andrew J. Davison\\
{\tt\small a.davison@imperial.ac.uk}\\
Department of Computing, Imperial College London, UK
}

\maketitle
\thispagestyle{empty}

\begin{abstract}
We discuss and predict the evolution of Simultaneous Localisation and Mapping
(SLAM) into a general geometric and semantic `Spatial AI' perception
capability for intelligent embodied devices. A big gap remains
between the visual perception performance that devices such as augmented reality eyewear or comsumer robots will require and
what is possible within the constraints imposed by real products.
Co-design of algorithms, processors and sensors will be
needed. We explore the computational structure of current and future
Spatial AI algorithms and consider this within the landscape of
ongoing hardware developments.
\end{abstract}


\section{Introduction}

While the usually stated goal of computer vision is to report `what' is `where' in an image in a general way,
{\em Spatial AI} is the online problem where vision is to be used,
usually alongside other sensors, as part of the Artificial
Intelligence (AI) which permits an embodied device to interact
usefully with its environment. This device must operate in real-time, in a context, and with
goals. It may be what we would
clasically think of as a robot, a fully autonomous artificial agent; 
or it could be a device like an AR headset designed to be used by a human to
augment their capabilities. As recently clarified by Markoff~\cite{Markoff:Book2016}, either
autonomous AI as in a robot or an `Intelligence Augmentation' (IA) system such as AR have very similar
requirements in terms of Spatial AI.

So the goal of a Spatial AI system is not abstract scene
understanding, but continuously to capture the right
information, and to build the right representations, to enable
real-time interpretation and action.  The design of such a system will
be framed at one end by task requirements on its performance, and at
the other end by constraints imposed by the setting of the device in
which it is to be used.

Let us consider the example of a mass market
household robot product of the future, which is set the tasks of
monitoring, cleaning and tidying a set of rooms. Its task requirements
will include the ability to check whether furniture and objects have
moved or changed; to clean surfaces and know when they are clean; to
recognise, move and manipulate objects; and to deal promptly and
respectfully with humans by moving out of the way or assisting
them. Meanwhile, its Spatial AI system, comprising one or more
cameras, supporting sensors,
processors and algorithms, will be constrained by factors including
the price, aesthetics, size, safety, and power usage, which must fit
within the range of a consumer product.

As a second example, this time from the IA domain, we imagine a future augmented reality system which should
provide its wearer with a robust spatial memory of all of the
places, objects and people they have encountered, enabling things such
as easily finding lost objects, and the placing of virtual notes or other
annotations on any world entity. To achieve wide adoption, the device should have the size,
weight and form factor of a standard pair of spectacles (65g), and operate
all day without needing a battery charge ($<$1W power usage).

There is currently a big gap between what such powerful Spatial AI
systems need to do in useful applications, and what can be
achieved with current technology under real world constraints. There
is much promising research on the algorithms and technology needed,
but robust performance is still difficult even when expensive, bulky
sensors and unlimited computing resources are
available. The gap between reality and desired performance
becomes much more significant still when the constraints on real products
are taken into account. In particular, the size, cost and power
requirements of  the computer processors currently enabling advanced robot
vision are very far from fitting the constraints imposed by these
applications we envisage.

The PAMELA project from the Universities of Manchester, Edinburgh and
Imperial College has opened up research in this area by taking a
broad look at the interaction of Spatial AI algorithms and the
increasingly heterogeneous processors they must run on, producing
pioneering work such as the SLAMBench
framework~\cite{Nardi:etal:ICRA2015,Bodin:etal:ICRA2018}.  As the project nears its end,
in this paper we take a long view of Spatial AI research, make some
analysis of the key computational structure of these problems, and
taking account of ongoing trends in processor technology make some
predictions for the future of this field.

\section{SLAM, Spatial AI and Machine Learning}

The research area of visual SLAM (Simultaneous Localisation and
Mapping) in robotics and computer vision has long been concerned with
real-time, incremental estimation of the shape and structure of the
scene around a robot, and the robot's position within it.  The level
of scene representation that has been possible in real-time SLAM has
gradually improved, from sparse features
(e.g.~\cite{Davison:ICCV2003}) to dense
(e.g.~\cite{Newcombe:etal:ICCV2011}) maps and now increasingly
semantic labels~\cite{McCormac:etal:ICRA2017,Tateno:etal:CVPR2017}.

Commercial SLAM provider SLAMcore Ltd.~\cite{SLAMcoreweb} refers to sparse
localisation, dense mapping and semantic labelling as Levels 1, 2 and 3
capabilities respectively. These are ongoing steps along SLAM's
evolution towards spatial perception. Observing the steady and
consistent progress of SLAM over more than 20 years, we have become
confident that the operation of current and forseeable SLAM systems is the best
guide we have for the algorithmic structure of future Spatial AI.

Spatial AI will be fundamental enabling technology for a the next
generation of smart robots, mobile devices and other products that we
can't yet imagine; a new layer of technology that could eventually be
pervasive.  In the recent technical report from leading UC Berkeley
researchers on systems challenges for AI
\cite{Stoica:etal:TechReport2017}, it is argued that devices which can
act intelligently in their environments via continual learning must be
capable of Simulated Reality (SR) which can ``faithfully simulate the
real-world environment, as the environment changes continually and
unexpectedly, and run faster than real time''.
Judea Pearl, recently discussing efficient situated learning and the
need to reason about causation~\cite{Pearl:TechReport2017}, argues that
``what humans possessed that other species lacked was a
mental representation, a blue-print of their environment which they
could manipulate at will to imagine alternative hypothetical
environments for planning and learning''. Both  of these describe
exactly what we envision as the end-goal of SLAM's ongoing evolution into Spatial AI.

Most Spatial AI systems will have multiple
applications, not all predictable at the time of design.
We therefore make the following hypothesis:
{\em
When a device must operate for an extended period of
time, carry out a wide variety of tasks (not all of which are
necessarily known at design time), and communicate with other entities
including humans, its Spatial AI system should build a general and persistent
scene representation which is close to metric 3D geometry, at least locally, and is human
understandable.
}
To be clear, this definition leaves a lot of space for many choices
about scene representation, with both learned and designed elements,
but rules out algorithms which make use of very specific task-focused
representations. 

Our  second hypothesis is that: {\em The usefulness of a
  Spatial AI system for a wide range of tasks is well represented by a
  relatively small number of performance measures.} That is to say that whether the system is to be used to guide the
autonomous flight of a delivery drone in a tight space, or a
household robot to tidy a room, or to enable an augmented reality
display to add synthetic objects to a scene, then even though these
applications certainly have different requirements and constraints,
the suitability of a Spatial AI system for each can be specified by
a the specification of a small number of performance parameters. The obvious parameters describe
aspects like global device localisation accuracy and update latency, but we
believe that there other metrics which will be more meaningful for applications, like
distance to surface contact prediction accuracy, object identification accuracy or tracking robustness.
We will discuss performance metrics further in Section~\ref{section:metrics}.

For the purposes of the rest of this paper, we will therefore call such a module
which incrementally builds and maintains a generally useful, close to
metric scene representation, in real-time and from primarily visual
input, and with quantifiable performance metrics, a {\em Spatial AI system}.

\subsection{Machine Learning}

In recent years, Machine Learning (ML) has increasingly come to the
fore in AI, and overcome human-designed approaches in many
problems. In machine learning, the parameters of a black box
computational unit which transforms inputs to outputs are learned by
adjusting them on the basis of training data, with the aim of
optimising its performance either when compared to explicit
labels provided by an external source (supervised learning), or more indirectly by judging performance against high level
task goals over a period of time (reinforcement learning). Machine
Learning contrasts with {\em estimation methods} where the
computational unit to achieve an AI task is explicitly hand-designed with
nameable variables, modules, algorithms and other structures.

Computer vision has proved to be a very
successful domain for the application of ML.
As is well known, Deep Learning approaches mainly
based
on Convolutional Neural Networks (CNNs) have become the
dominant approach to achieving state of the art results in many vision
problems, such as image classification or segmentation.

More
recently, deep
learning has started to show promising results on
problems in vision-guided robotics (e.g. \cite{Finn:etal:ARXIV2015}).
In these approaches, a deep network which processes the
raw pixels of incoming images is trained `end to end' by supervision from a
non-learned system (e.g.~\cite{James:etal:CORL2017}), or via reinforcement based on the
the achievement of discrete tasks such as object placement
or local navigation.
These networks learn directly to output motor control signals based on
visual input, and therefore any internal representation they need
(such as the shape of the environment, and the robot's
position within it) are represented within the network itself as
required to achieve the task, in an implicit form which is not accessible beyond
the task at hand.

In general Spatial AI problems, there has so far been less success
in machine learning methods which incrementally improve a world
representation over time from multiple measurements.
Such learning requires a computational unit which has memory and
captures its own internal representaion over time. A Recurrent Neural Network (RNN) is the
fundamental concept of a network whose activations and outputs depend
not just `single shot' on the current input data 
but also on internal states which are the outcome of
previous inputs. There are many related ideas and also efforts to
interface deep networks with explicit memory blocks.

In Spatial AI, training an RNN or similar to produce useful output
sequentially from a real-time stream of input data requires it to
capture within its internal state a persistent set of concepts which
must closely relate to the shape and qualities of the environment
around the device.  We have certainly seen some success with methods
aiming to do this, such as the work of Wen~\etal~\cite{Wen:etal:ICRA2017}
on estimating incremental visual odometry from video using an RNN
whose training was via a supervised known pose signal. 

However, a group of methods which seems very promising aims to impose
structure on what is learned by a network. 
Gupta~\etal~\cite{Gupta:etal:ARXIV2017} presented a method for local
navigation which forces a deep network to learn
about a robot's environment in a manner which tends towards a metric grid
by presenting it with metric
spatial transformations based on the robot's known motion in 2D.
Zhou~\etal~\cite{Zhou:etal:CVPR2017} have excitingly shown how networks to estimate incremental camera motion and scene depth can be trained in a coupled manner from unlabelled video data, using photoconsistency via the geometric warping between textured depth frames as self-supervision.

These are learning architectures which use the designer's knowledge of
the structure of the underlying estimation problem to increase what
can be gained from training data, and can be seen as hybrids between
pure black box learning and hand-built estimation.  Another key paper
in this area introduced the idea of Spatial Transformer
Networks~\cite{Jaderberg:etal:ARXIV2015}, as generalised in
Handa~\etal's gvnn~\cite{Handa:etal:ARXIV2016}.  Why should a neural
network have to use some of its capacity to learn a well understood
and modelled concept like 3D geometry? Instead it can focus its
learning on the
harder to capture factors such as correspondence of surfaces under
varying lighting.

These insights support our first hypothesis that future Spatial AI
systems will have recognisable and general mapping capability which
builds a close to metric 3D model. This `SLAM' element may either be
embedded within a specially architected neural network, or be more
explicitly separated from machine learning in another module which
stores and updates map properties using standard estimation methods
(as in SemanticFusion~\cite{McCormac:etal:ICRA2017} for instance).
In both cases, there should be an identifiable region of memory which contains a representation of space with some recognisable geometric structure.

Besides utility, such AI systems will have the additional advantage of
using representations that can be communicated to and understood by
humans, and therefore in principle controlled.

\subsection{Closed Loop SLAM}

As a sensor platform carrying at least one camera and other sensors moves through the world, its motion either under active control or provided by another agent,
the essential algorithmic ways that a Spatial AI system of any
variety works can be summarised as follows:
\be
\item
  Starting from and continuing to make use of prior knowledge of the type of scene it is working in, the system uses data from its sensors to incrementally build and refine a persistent world model
  which captures abstracted geometric, appearance and semantic
  information  about its environment. It also models and incrementally estimates
  the state of the moving camera platform relative to the world.
The amount of data stored to represent a region of space of a certain size will have a maximum bound so that if the sensor spends a long time in one region the size of the representation does not grow arbitrarily.
\item As new image data is captured from the moving camera, it is
  compared with the current world model. With low latency, the system 
must decide which new data is not important and which can be used to
update the model's persistent estimates. The task of matching up
sensor data to the relevant parts of the model is called {\em data association}.
\item The model update is often considered as consisting of {\em
  tracking}, which is getting new estimates of the sensor platform's
  position/state, and {\em map update} where the scene model is
  improved and expanded. This division is however somewhat arbitrary.
\ee

The key computational quality of this approach is its {\em closed
  loop} nature, where the world model is persistent and incrementally
updated, representing in an abstracted form all of the useful data which has
been acquired to date, and is used in the real-time loop for data association and
tracking. This is in contrast with vision systems which perform
incremental estimation (such as visual odometry, where camera motion
is estimated from frame to frame but long term data structures are not
retained), or which can
only achieve global consistency with off-line, after the event
batch computation. That is not to say that in a closed loop Spatial AI
system every computation should happen at a fixed rate, but
more importantly that it should be available when needed to allow
real-time operation of the whole system to continue without pauses.

Closed loop SLAM has generally been enabled by probabilistic estimation, the fundamental approach building models which summarise past data in a form which acknowledges
uncertainty and allows new data to be correctly weighted and fused.
In sparse feature-based SLAM, probabilistic fusion is implemented via
tools such as the Extended Kalman Filter (as
in~\cite{Davison:ICCV2003}) or incremental non-linear optimisation (as in~\cite{Leutenegger:etal:IJRR2014}).

Sparse feature maps are useful as landmark sets for position
estimation, but do not provide much information about the scene around
a camera, and progress has more recently been towards systems aiming
to do better, via dense mapping and now semantic mapping. Ultimately,
if a scene model is fully {\em generative} then it can be used to make
complete predictions about sensor data. This is important because then
in principle {\em every} piece of sensor data can be compared with the
model prediction to uncover what is previously unseen or changed in
the scene. In simple terms, for a system and to recognise something
which is unusual, it must keep up to date a full predictive model of
what is normal.


The latest real-time SLAM systems are hybrids, combining 
estimation of tracking and dense surface shape with learned
components for labelling and recognition
(e.g. SemanticFusion~\cite{McCormac:etal:ICRA2017}).
As mentioned
earlier, other learned components are now looking promising for other
elements of a full system such as VO and depth estimation
(e.g. Zhou~\etal~\cite{Zhou:etal:CVPR2017}). We forsee these
components coming together in an increasingly tight and interlinked
fashion, helping  and feeding into each other. 
While SemanticFusion currently uses geometric SLAM and CNN-based
labelling as essentially separate modules which are fused into a final
labelled map
output, systems coming in the near future will be closely integrated
in a tight loop which will surely give much better performance. For
instance, a CNN for labelling which takes as input not just a new
live image frame but also the current set of label distributions
reprojected from the 3D map should be much better, and ideas like this
are starting to be proven~\cite{Xiang:Fox:ARXIV2017}.

One clear goal is to create a SLAM system which can as
far as possible understand a scene directly and efficiently at the
level of recognised objects and entities (as in the prototype system
SLAM++~\cite{Salas-Moreno:etal:CVPR2013}), reserving bottom-up
reconstruction and labelling for unfamiliar elements. Whether the components
making up such a system are designed or learned may not have a big
impact on the overall computational structure.

Our main aim in the rest of this paper is therefore to analyse the computational structure of
Spatial AI systems, while avoiding pinning down specifics where this
is difficult, and to make some predictions and tentative
design choices about their implementation going forward.
A crucial element of this is that Spatial AI is fundamentally an
embodied, real-time problem. We believe strongly that the design of
the algorithms and data structures required should take place in a
tightly integrated manner with that of the physical processors and sensors which
together form a whole system. In the next section we will therefore
consider the relevant landscape in processor and sensor design.

\section{The Future Landscape of Processor and Sensor Hardware}

SLAM research was for many years conducted in the era when single core
CPU processors could reliably be counted on to double in clock speed,
and therefore serial processor capability, every 1--2 years. In
recent years this has stopped being true. The strict definition of
Moore's Law, describing the rate of doubling of transistor density in
integrated circuits, has continued to hold well into the current
era. What has changed is that this can no longer be proportionally
translated into serial CPU performance, due to the breakdown of
another less well known rule of thumb called Dennard Scaling, which
states that as transistors get smaller their power density stays
constant. When transistors are reduced down to today's nanometre
sizes, not so far from the size of the atoms they are made from, they
leak current and heat up. This `power wall' limits the clock
speed at which they can reasonably be run without overheating
uncontrollably to something around 4GHz.

Processor designers must therefore increasingly look towards
alternative means than simply faster clocks to improve computation
performance. The processor landscape is becoming much more complex,
parallel and specialised, as described well in Sutter's online article
`Welcome to the Jungle'
\cite{Sutter:Jungle2011}.
Processor design is becoming more varied and complex even in `cloud'
data centres. Pressure to move away from CPUs is even stronger in
embedded applications like Spatial AI, because here power usage is a
critical issue, and parallel, heteregeneous, specialised processors
seem to be the only route to achieving the computational performance
Spatial AI needs within power restrictions which will fit real
products.
So while current embedded vision systems, e.g. for drones, often use CPU
only implementations of SLAM algorithms (rather than requiring GPUs), we believe that this is not the right
approach for the longer term. While current desktop GPUs are certainly
power-hungry, Spatial AI must eventually
embrace parallelism and heterogeneity in computation, and accept
that this will be  the dominant paradigm going forward for practical systems.

Mainstream geometrical computer vision started to take advantage of parallel
processing in the form of GPUs nearly 10 years ago
(e.g. \cite{Pock2007:etal:CVWW2007}), and in Spatial AI this led to
breakthroughs in dense
SLAM~\cite{Newcombe:etal:ICCV2011,Newcombe:etal:ISMAR2011}. The SIMT (Single Instruction, Multiple Threads)
parallelism that GPUs provide is well suited to elements of real-time
vision where the same operation needs to be applied to every element
of a regular array in image or map space. 
Concurrently, GPUs were central to the emergence of deep learning in
computer vision~\cite{Krizhevsky:etal:NIPS2012}, by providing the
computational resource to enable neural
networks of sufficient scale to be trained to finally prove their
worth in significant tasks such as image classification.

The move from CPUs to GPUs as the main processing workhorse for
computer vision is only the beginning of how processing technology is
going to evolve. We forsee a future where an embedded Spatial AI
system will have a heterogeneous, multi-element, specialised
architecture, where low power operation must be achieved together with
high performance. A standard SoC (system-on-chip) for embedded vision
ten years from now, which might be used in a personal mobile device,
drone or AR headset, will be likely to still have elements which
are similar in design to today's CPUs and GPUs, due to their
flexibility and the huge amount of useful software they can
run. However, it is also likely to have a number of specialised
processors optimised for low power real-time vision.

The key to efficient processing which is both fast and consumes little
power is  {\em to divide computation between a large number of relatively
low clock-rate or otherwise simple cores, and
to minimise the movement of data between them}.
A CPU pulls and pushes
small pieces of data one by one from and to a separate main memory
store as it performs computation, with local caching of regularly used
data the only mechanism for reducing the flow.
Programming for a single CPU is straightforward, because any 
type of algorithm can be broken 
down into sequential steps with access to a single central memory
store, but the piece by piece flow of data to and from central memory
has a huge power cost.

More efficient processor designs aim to keep processing and the data
operated on close together, and to limit the transmission of intermediate results. The ideal way to achieve this is a close match
between the design of a processor/storage architecture and the
algorithm it must run. A GPU certainly has large advantages over a CPU
for many computer vision processing tasks, but in the end a GPU is a
processor designed originally for computer graphics rather than vision
and AI. Its SIMT architecture can
efficiently run algorithms where the same operation is carried out
simultaneously on many different data elements. In a full Spatial AI
system, there are still many aspects which do not fit well with this,
and a joint CPU/GPU architecture is currently needed with substantial data
transfer between the two.

While it is relatively accessible to design custom `accelerator'
processors which could implement certain specific low-level algorithms
with high efficiency (see for instance the OpenVX project from the
Khronos Group), there has been relatively little work until recently
on thinking about the whole computational structure of whole close
loop embedded systems like Spatial AI.  It is certainly true now
that low power vision is seen as an increasingly important aim in
industry, and custom processors to achieve this have been developed
such as Movidius' Myriad series. These processors combine low power
CPU-like, DSP-like and custom elements in a complete package.
The `HPU' custom-designed by Microsoft for their Hololens AR headset is rather similar in design, and the recently announced second version will include additional custom hardware support for neural networks.

If we try to look further ahead, we can conceive of processor designs
which offer the possibility of a much closer match between
architecture and algorithms. Highly relevant to our aims are major
efforts which are now taking place on new ways of doing large-scale
processing by being made up from large numbers of independent and
relatively low-spec cores with the emphasis on
communication. SpiNNaker~\cite{Furber:etal:IEEE2014} is a major research project from the
University of Manchester which aims to build machines to emulate
biological brains.  It has produced a prototype machine made up from
boards which each have hundreds of ARM cores, and with up to a million
cores in total. With the rather different commercial aim of providing
an important new type of processor for AI, Graphcore is a UK
company developing `IPU' processors which comprise thousands of highly
interconnected cores on a single chip.

Both of these projects are primarily being designed with efficient implementation of
neural networks in mind, in the case of both SpiNNaker and Graphcore
with the belief that the important matter is the overall topology of a
large number of cores, each performing different operations but highly
and efficiently inter-connected in a graph configuration adapted to
the use case. These designs have not taken strong decisions about the
type of processing carried out at each core, or the type of messages
they can exchange, with the desire to leave these matters to the
choice of future programmers. This is as opposed to more explicitly
neuromorphic architectures aiming to implement particular models of
the operation of biological brains, such as IBM's Truenorth project.

Such architectures are not yet close to ready for embedded vision, but seem to offer great long term potential for the design of
Spatial AI systems where the graph structure of the algorithms and
memory stores we
use can be matched to the implementation on the processor in a custom
and potentially highly efficient way. We will consider this in more detail later on.

But we also believe that we should go further than thinking of mapping
Spatial AI to a single processor, even when it has an internal graph
architecture. A more general concept of a graph applies to
communication to cameras and other sensors, actuators and other
outputs, and potentially entirely off-board computing resources in the
cloud. 

A particularly important consideration is the real-time data flow between the
one or more camera sensors in a Spatial AI system and the main
processing resource where map storage and processing occurs. Video
data is huge and expensive to transmit. However, it is highly
redundant, both temporally and spatially: nearby pixels in both space
and time tend to have similar values.

The concept of a camera is today becoming
increasingly broad with ongoing innovation by sensor designers, and
for our purposes we consider any device which essentially captures an
array of light measurements to be a visual sensor. Most are
passive in that they record and measure the ambient light which
reaches them from their surroundings, while another large class of
cameras emit their own light in a more or less controlled
fashion.
In Spatial AI, many types of camera have been used, with the most
common being passive monocular and stereo camera rigs, and depth 
 cameras based on structured light or time of flight
concepts.
Every camera design represents a choice in terms of the quality of
information it provides
(measured in such ways as spatial and temporal resolution and dynamic
range), and the constraints it places on the system it is used in such
as size and power usage. In previous work~\cite{Handa:etal:ECCV2012} we studied some of the trade-offs possible between performance and computational cost in the Spatial AI sub-problem of real-time tracking.

One type of visual sensor which is particularly
promising for Spatial AI is known as the event camera. This is a
sensor which instead of capturing a sequence of full image frames as a
standard video camera does, only records and transmits changes in
intensity. The output of the
sensor is a stream of `events' from the independently sensitive
analogue pixels, each with pixel coordinates and an accurate
timestamp. The principle is that the event stream encodes the content of video at a much
lower bit-rate, while offering advantages in time temporal and
intensity sensitivity, and it has recently been demonstrated
(e.g.~\cite{Kim:etal:ECCV2016}) that SLAM algorithms can be formulated
which estimate camera motion, scene intensity and depth from only the
event stream.

The event camera is surely only the starting point for coming rapid changes
in image sensor technology, where low power computer vision will be an
increasingly important driver. We can expect a full generalisation of the
concept of the event camera to sensors which perform significant
processing of intensity data as part of the sensing process itself,
and will communicate with a main processor in a bidirectional manner
in an abstracted form which is very different from sensing raw video streams.

\begin{figure}[t]
\centerline{
\hfill
\includegraphics[width=1.0\columnwidth]{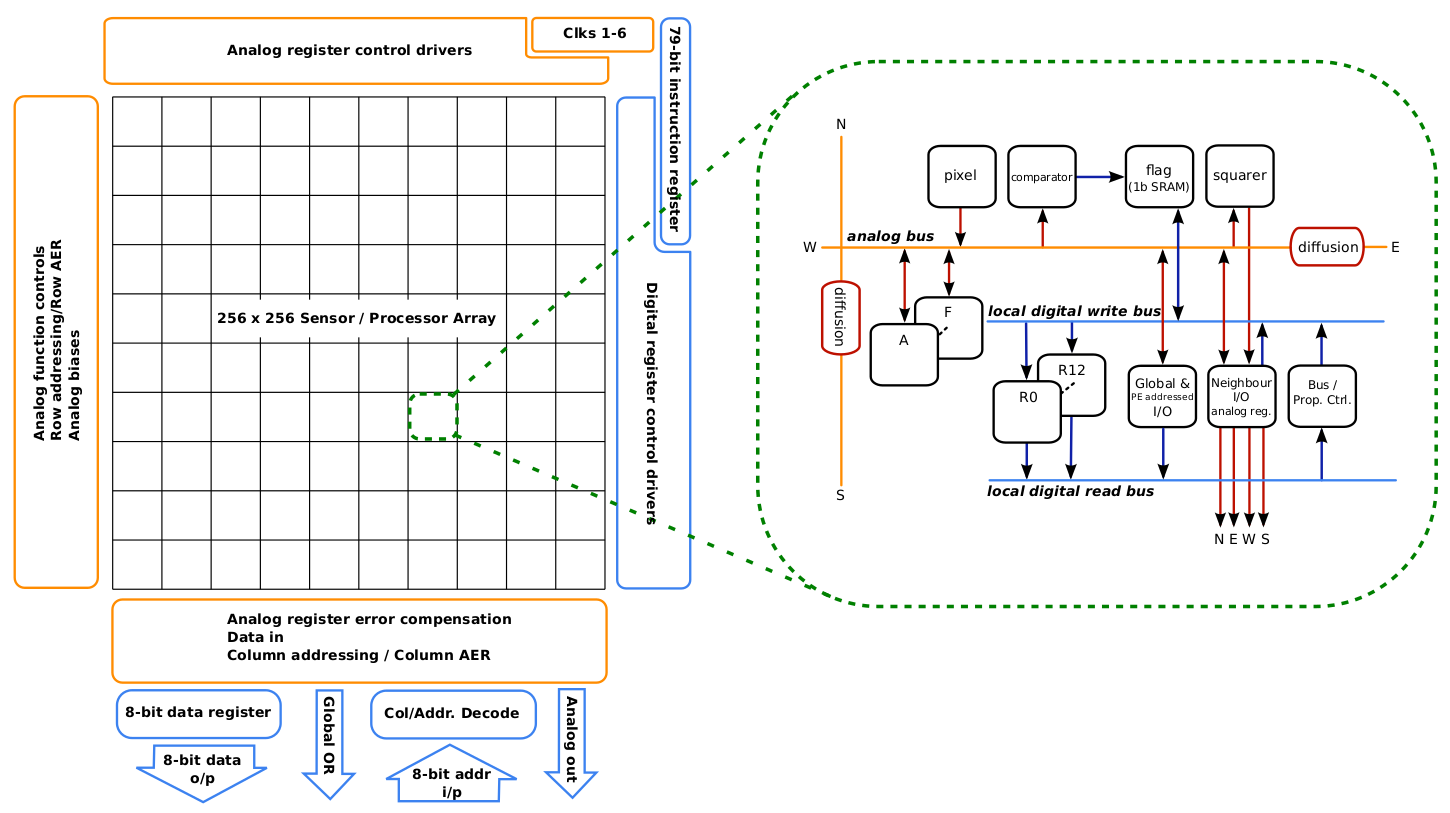}
\hfill
}
\caption{\label{fig:scamp5}
The SCAMP5 architecture for integrated visual sensing and processing
(Figure taken from Martel and
Dudek~\cite{Martel:etal:ASRMOV2016}, and reproduced courtesy of
the authors.)
}
\vspace{2mm} \hrule
\end{figure}

One significant ongoing academic project in this space is the SCAMP
series of vision chips with in-plane processing from the University of
Manchester (see \cite{Martel:etal:ASRMOV2016} for an introduction),
and Figure~\ref{fig:scamp5} taken from that paper.
The SCAMP5 chip runs at 1.2W and has an image resolution of 
256$\times$256, with each pixel controlled by and processable by
per-pixel processing. Using analog current-mode
circuits, summation, subtraction, division, squaring, and
communication of values with neighbouring pixels can be achieved
extremely rapidly and efficiently to permit a significant level of  real-time vision
processing completely on-chip. Ultra-low power operation can
alternatively already be
achieved in applications where low update rates are sufficient.

As we will discuss later on, the range of vision processing which
could eventually be performed by such an image plane processor is
still be to fully discovered. The obvious use is in front-end
pre-processing such as feature detection or local motion tracking. We
believe that the longer term potential is that while a central
processor will be required for full model-based Spatial AI,
close-to-sensor processing can interact fully with this via two-way
low communication, with the main aim of reducing the bit-rate needed
between the sensor and main processor and therefore the communication
power requirements.

Finally, when considering the evolution of the computing resources for
Spatial AI, we should never forget that, cloud computing resources
will continue to expand in capacity and reduce in cost. All future
Spatial AI systems will likely be cloud-connected most of the time,
and from their point of view the processing and memory resources of
the cloud can be assumed to be close to infinite and free. What is not
free is communication between an embedded device and the cloud, which
can be expensive in power terms, particularly if high bandwidth data
such as video is transmitted. The other important consideration is the
time delay, typically of significant fractions of a second, in sending
data to the cloud for processing.

The long term potential for cloud-connected Spatial AI is clearly enormous.
The vision of Richard Newcombe, Director of Research Science at Oculus, 
is that all of these devices should communicate and collaborate to
build and maintain shared a `machine perception map' of the whole
world. The master map will be stored in the cloud, and individual
devices will interact with parts of it as needed. A
shared map can be much more complete and detailed than that build by
any individual device, due to both sensor coverage and the computing
resources which can be put into it. A particularly interesting point
is that the Spatial AI work which each individual device needs to do
in this setup can in theory be much reduced. Having estimated its
location within the global map, it would not need to densely map or
semantically label its surrounding if other devices had already done
that job and their maps could simply be projected into its field of
view. It would only need to be alert for changes, and in turn play its
part in returning updates.

\section{High Level Design}

We now turn more specifically to a design for the architecture of a
Spatial AI device.  Despite the clear potential for cloud-connected
shared mapping, here we choose to focus purely on a single device
which needs to operate in a space with only on-board resources,
because this is the most generally capable setup which could be useful
in any application and not require additional infrastructure.

The first thing to consider in the design of our hypothetical future
Spatial AI system is what it will be required to do:
\be
\item
Our system will comprise one or more cameras, and supporting
  sensors such as an IMU, closely integrated with a processing
  architecture in  a small, low power package which is embedded in a
  mobile entity such as a robot or AR system. While much can be
  achieved in SLAM and vision with a single camera as the only sensor,
  it is clear that most practical applications will observe their
  surroundings with multiple cameras and support this with other
  appropriate sensors. For instance, a future household robot is
  likely to have navigation cameras which are centrally located on its
  body and specialised extra cameras, perhaps mounted on its wrists to
  aid manipulation. 
\item
In real-time, the system must maintain and update a world model, with
  geometric and semantic information, and estimate its position within that model, from primarily or only measurements from its on-board sensors.
\item
The system should provide a wide variety of task-useful information
about `what' is `where' in the scene. Ideally, it will provide a full
semantic level model of the identities, positions, shapes and motion
of all of the objects and other entities in the surroundings.
\item
  The representation
  of the world model will be close to metric, at least locally, to
  enable rapid reasoning about arbitrary predictions and measurements
  of interest to an AI or IA system.
\item
  It will probably retain a maximum quality representation of geometry
  and semantics only in a focused manner; most obviously for the part
  of the scene currently observed and relevant for near-future
  interaction. The rest of the model will be stored at a hierarchy of
  residual quality levels, which can be rapidly upgraded when
  revisited. 
\item
The system will be generally alert, in the sense that
  every incoming piece of visual data is checked against a forward
  predictive scene model:  for tracking, and for detecting changes in the environment  and independent
  motion. The system will be able to respond to changes in its environment.
\ee
The next element of our high level thinking is to identify the core ways that we can achieve all of this with high performance but low power requirements.

We believe that the key to efficient processing in Spatial AI is to
identify the graphs of computation and data movement in the algorithms
required, and as far as possible to make use of or design processing
hardware which has the same properties, with the particular goal of
minimising data movement around the architecture. We need to identify the following things:
What is stored where? What is processed where? What is transmitted where, and when?

We will not attempt here to draw strong parallels with neuroscience,
but clearly there is much scope for relating the ideas and designs we
discuss here in artificial Spatial AI systems with the vision and
spatial reasoning capabilities and structures of biological brains.
The human brain apparently achieves high performance, fully `embedded'
semantic and geometric vision using less than 10 Watts of power, and
certainly its structures have properties which mirror some of the
concepts we discuss. We will leave it to other authors to analyse the
relationships further; this is mostly due to our lack of expertise in
neuroscience, but also partly due to a belief that while artificial
vision systems clearly still have a great deal to learn from biology,
they need not be designed to replicate the performance or structure of
brains. We we consider the Spatial AI computation problem purely
from the engineering point of view, with the goal of achieving the
performance we need for applications while minimising resources. It
should surely not be surprising that some aspects of our solutions
should mimic those discovered by biological evolution, while in other
respects we might find quite different methods due to two contrasts:
firstly between the
incremental `has to work all of the time' design route of evolution
and the increased freedom we have in AI design; and secondly between the
wetware and hardware available as a computational substrate.

\section{Graphs in SLAM}

\label{section:graphs}

SLAM and the wider Spatial AI problem have some immediately apparent
graph structures built into them. We will here aim to identify and
discuss these.

\subsection{Geometrical Graphs}

\label{section:geomgraphs}

There are well known
graphs which emerge naturally from the geometry of
cameras and 3D spaces and the data which represents these.

\subsubsection{Image Graph}


First, there is the regular, usually
rectangular geometry of the pixels which make up the image sensor of a
camera. While each of these pixels is normally independently sensitive
to light intensity, many vision algorithms make use of the fact that
the output of nearby pixels tends to be strongly correlated. This is because nearby
pixels often observe parts of the same scene objects and
structures. Most commonly, a regular graph in which each pixel is
connected to its four (up, down, left, right) neighbours is used as
the basis for smoothing operations in many early vision problems such as dense matching or optical flow estimation (e.g.~\cite{Pock:PHD2008}).

The regular graph structure of images is also taken advantage of by
the early convolutional layers of CNNs for all kinds of computer
vision tasks.  Here the multiple levels of convolutions also acknowledge
the typical hierarchical nature of local regularity in image data.

\subsubsection{Map Graph}

\label{sec:mapgraph}

The other clear geometrical graph present in Spatial AI problems is in the
maps or models which are built and maintained of an environment by
SLAM systems.
This graph structure has been recognized and made use of by many important SLAM methods (e.g. \cite{Kaess:etal:IJRR2012,Konolige:Agrawal:TRO2008,Salas-Moreno:etal:CVPR2013,Engel:etal:ECCV2014,Mei:etal:IJCV2011}).

As a camera moves through and observes the world, a SLAM algorithm
detects, tracks and inserts into its map features which are extracted
from the image data. (Note that we use the word `feature' here in a
general sense to mean some abstraction of a scene entity, and that we
are not confining our thinking to sparse point-like landmarks.)
Each feature in the scene has a region of camera positions from which
it is measurable. A feature will not be measurable if it is outside of
the camera's field of view; if it is occluded; or for other reasons
such that its distance from the camera or angle of observation are very different from when it was first observed.

This means that as the camera moves through a scene, 
features become observable in variable overlapping patterns. As
measurements are made of the currently visible features, estimates of
their locations are improved, and the measurements are also used to
estimate the camera's motion. The estimates of the locations of
features which are observed at the same time become strongly
correlated with each other via the uncertain camera state. Features
which are `nearby' in terms of the amount of camera motion between
observing them are still correlated but somewhat less strongly; and
features which are `distant' in that a lot of motion (and
SLAM based on intermediate features) happens between their observation
are only weakly correlated.

The probabilistic joint density over feature locations which is the
output of SLAM algorithms can be efficiently represented by a graph
where `nearby' features are joined by strong edges, and `distant' ones
by weak edges. A threshold can be chosen on the accuracy of
probabilistic representation which leads to the cutting of weaker
edges, and therefore a sparse graph where only `nearby' features are
joined.

One way to do SLAM is not to explicitly estimate the state of scene
features, but instead to construct a map of a subset of the historical
poses that the moving camera has been in, and to keep the scene map
implicit. This is usually called pose graph SLAM, and within this kind
of map the graph structure is obvious because we join together poses
between which we have been able to get sensor correspondence. Poses
which are consecutive in time are joined; and poses where we are able
to detect a revisit after a longer period of time are also joined
(this is called a loop closure).  Whether the graph is of historic
poses, or of scene features, its structure is very similar, in that it
connects either `nearby' poses or the features measured from those
poses, and there is not a fundemental difference between the two
approaches.

A property of a passive camera which is different from other sensors is
that it has effectively infinite range.
It can see objects which are very close in exquisite detail; but also
observe objects which are extremely far away. Visual maps will consist
of all of these things, and this is why we have not been precise so
far with the concepts of `nearby' and `distant'. The locality in
visual maps is not a matter of simple metric distance. Remembering
that what is important is how feature position estimates become
correlated due to camera motion and measurements, a SLAM graph will
have a multi-scale character, such that elements measured from a close
camera (the different objects on a table) may form a significant
interconnected chunk of a graph, while another similar chunk contains much
more separated elements seen from farther away (a group of buildings
on the other side of the city). 

This leads us to conclude that
{\em
the most likely representation for Spatial AI is  to represent
3D space by a graph of features, which are linked in multi-scale
patterns relating to camera motion and together are able of generating
dense scene predictions.
}

Chli and Davison~\cite{Chli:Davison:ICRA2009} investigated an
automatic way to discover the hierchical graph structure of a
feature-based map by analysing correlations purely in measurement
space, and it is clear that relating such a structure quite closely to
a co-visibility keyframe graph gives a similar result. However, this
work was based on standard point features. We have not yet discovered
a suitable feature representation which describes both local
appearance and geometry in such a way that a relatively sparse feature
set can provide a dense scene prediction. We believe that learned
features arising from ongoing geometric deep learning research will
provide the path towards this.

Some very promising recent work which we believe is heading in the
right direction Bloesch~\etal's
CodeSLAM~\cite{Bloesch:etal:CVPR2018}. This method uses an image-conditioned
autoencoder to discover an optimisable code with a small number of
parameters which describes the dense depth map at a keyframe. In
SLAM,  camera poses and these depth codes can be jointly optimised to
estimate dense scene shape which is represented by relatively few
parameters. In this method, the scene geometry is still locked to
keyframes, but we believe that the next step is to discover
learned codes which can efficiently represent both appearance and 3D
shape, and to make these the elements of a graph SLAM system.


\subsubsection{Image/Model Correspondence}

Let us go a little deeper into this idea, and in particular to understand that 
as a camera moves through the world, the
contact or {\em correspondence} between its image graph and its map graph
will change, continuously and dynamically, in a way which can be very
rapid but is not random. What does this mean for the computational
structures we should use in Spatial AI?

Fundamentally, in a SLAM process we cannot pre-compute the
shape of the map that will be built, because it depends on the 
motion of the camera. However,  perhaps we can have a generic
structure which is representative of many types of motion, ready to be
filled when needed as a space is explored. What is important in graph maps, from a computational point of view, is {\em topology}: which things are connected to which others, and in which patterns.
Maps with hierarchical topologies in regular space such as multi-scale
occupancy grids or heightmaps have been developed (e.g.
\cite{Wurm:etal:ICRAW2010,Zienkiewicz:etal:3DV2016}), and these enable efficient mapping in systems where localisation can be separated and assumed to be good. In general visual SLAM, the graph needs a flexible structure, which is multiscale but can also be adapted at loop closure, and there are still good questions about how to design the ideal container.
Experiments could be performed with current dense SLAM systems to
determine the typical patterns of correlation between surface elements
in various applications.

\subsection{Computation Graph}

\label{section:computegraphs}

The other key type of graph which we can easily identify in Spatial AI problems is in the computation which takes place in a SLAM
system's real-time loop.

\begin{figure*}[t]
\centerline{
\hfill
\includegraphics[width=1.0\textwidth]{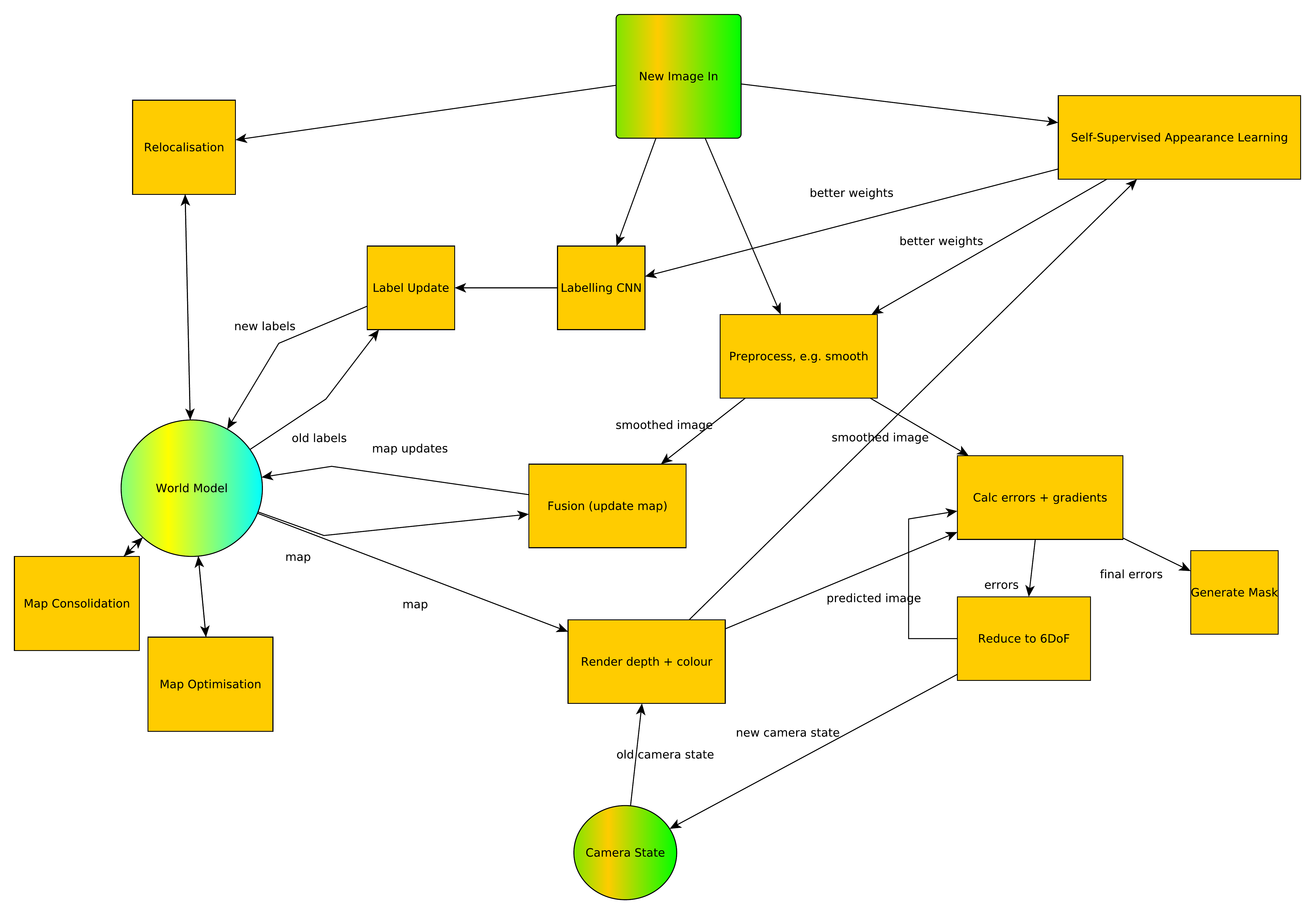}
\hfill
}
\caption{\label{fig:rvgraph}
Some elements of the Spatial AI real-time computation graph.
}
\vspace{2mm} \hrule
\end{figure*}

We make a hypothesis that the core computation graph for the tightest
real-time loop of future SLAM systems will have many elements which
are familiar with today's systems. Specifically the Dense SLAM
paradigm introduced by
Newcombe~\etal~\cite{Newcombe:PHD2012,Newcombe:etal:ISMAR2011,Newcombe:etal:ICCV2011}
is at the centre of this in our view, because this approach aims to
model the world in dense, generative detail such that every new pixel
of data from a camera can be compared against a model-based
prediction. This allows systems which are generally `aware', since as
they continually model the state of the world in front of the camera,
they can detect when something is out of place with respect to this
model, and therefore dense SLAM systems are now being developed which are moving beyond static scenes to reconstruct and track dynamic scenes~\cite{Newcombe:etal:CVPR2015,Runz:Agapito:ICRA2017}.
Dense SLAM systems can also make the best possible use of
scene priors, which will increasingly come from learning rather than being
hand-designed. 

Following some of these arguments, Newcombe~\etal's KinectFusion
algorithm~\cite{Newcombe:etal:ISMAR2011} was chosen as the basis for
the first version of the SLAMBench
framework~\cite{Nardi:etal:ICRA2015} which aimed to provide a forward
looking benchmark for the computational properties of SLAM systems.
Since KinectFusion, dense SLAM has been augmented with semantic
labelling in systems like SemanticFusion
\cite{McCormac:etal:ICRA2017}, which uses the surfel-based and loop
closure-capable ElasticFusion~\cite{Whelan:etal:RSS2015} as its dense
SLAM basis. As well as being used for SLAM, incoming image and depth
frames are fed to a CNN which has been pre-trained for per-pixel
semantic labelling into thirteen classes typical of domestic scenes,
such as wall, floor, table, chair or bed. The output for each pixel is
a distribution over possible labels, and these are then projected onto
the 3D SLAM map, where each surfel maintains and updates a label
probability distribution via Bayesian fusion.

As we discussed before, a goal of this line of research is to get to a
general SLAM system which has the ability to identify and estimate the
locations of all of the objects in a scene, in the style demonstrated
in a prototype way by SLAM++ \cite{Salas-Moreno:etal:CVPR2013} which made efficient maps directly at the level of objects in precise 3D poses, but
could only deal with a small number of specific
object types. How can we get back to this `object-oriented SLAM'
capability in the much more general sense, where a wide range of
object classes of varying style and details could be dealt with?
As discussed before,
SLAM maps of the future will probably be represented as multi-scale
graphs of learned features which describe geometry, appearance and
semantics. Some of these features will represent immediately recognised whole
objects as in SLAM++. Others will represent generic semantic elements
or geometric parts (planes, corners, legs, lids?) which are part of
objects either already known or yet to be discovered. Others may
approach surfels or other standard dense geometric elements in
representing the geometry and appearance of pieces whose
semantic identity is not yet known, or does not need to be known.

Recognition, and unsupervised learning, will operate on these feature
maps to cluster, label and segment them.
The machine learning methods which do this job will themselves
improve by self-supervision during the SLAM process, taking advantage of dense SLAM's properties as a `correspondence engine' \cite{Schmidt:etal:ICRA2017}.

From a starting point of the algorithmic analysis of the KinectFusion
algorithm in SLAMBench~\cite{Nardi:etal:ICRA2015}, we make an attempt at drawing a computation graph for a generally
capable future Spatial AI system in Figure~\ref{fig:rvgraph}.

Most computation relates to the {\em world model}, which is a
persistent, continuously changing and improving data store where the
system's generative representation of the important elements of the scene is
held; and the input {\em camera data stream}. Some of the main
computational elements are:
\bi
\item Empirical labelling of images to features (e.g. via a CNN).
\item Rendering: getting a dense prediction from the world map to
  image space.
\item Tracking: aligning a prediction with new image data, including
  finding outliers and detecting independent movement.
\item Fusion: fusing updated geometry and labels back into the map.
\item Map consolidation: fusing elements into objects, or imposing smoothing, regularisation.
\item Relocalisation/loop closure detection: detecting self similarity
  in the map.
\item Map consistency optimisation, for instance after confirming a loop closure.
\item Self-supervised learning of correspondence information from the
  running system.
\ei
In the following sections we will think about where these data and
computational elements might operate in future architectures.

\section{Main Map Processing; Representation, Prediction and Update}

\begin{figure}[t]
\centerline{
\hfill
\includegraphics[width=1.0\columnwidth]{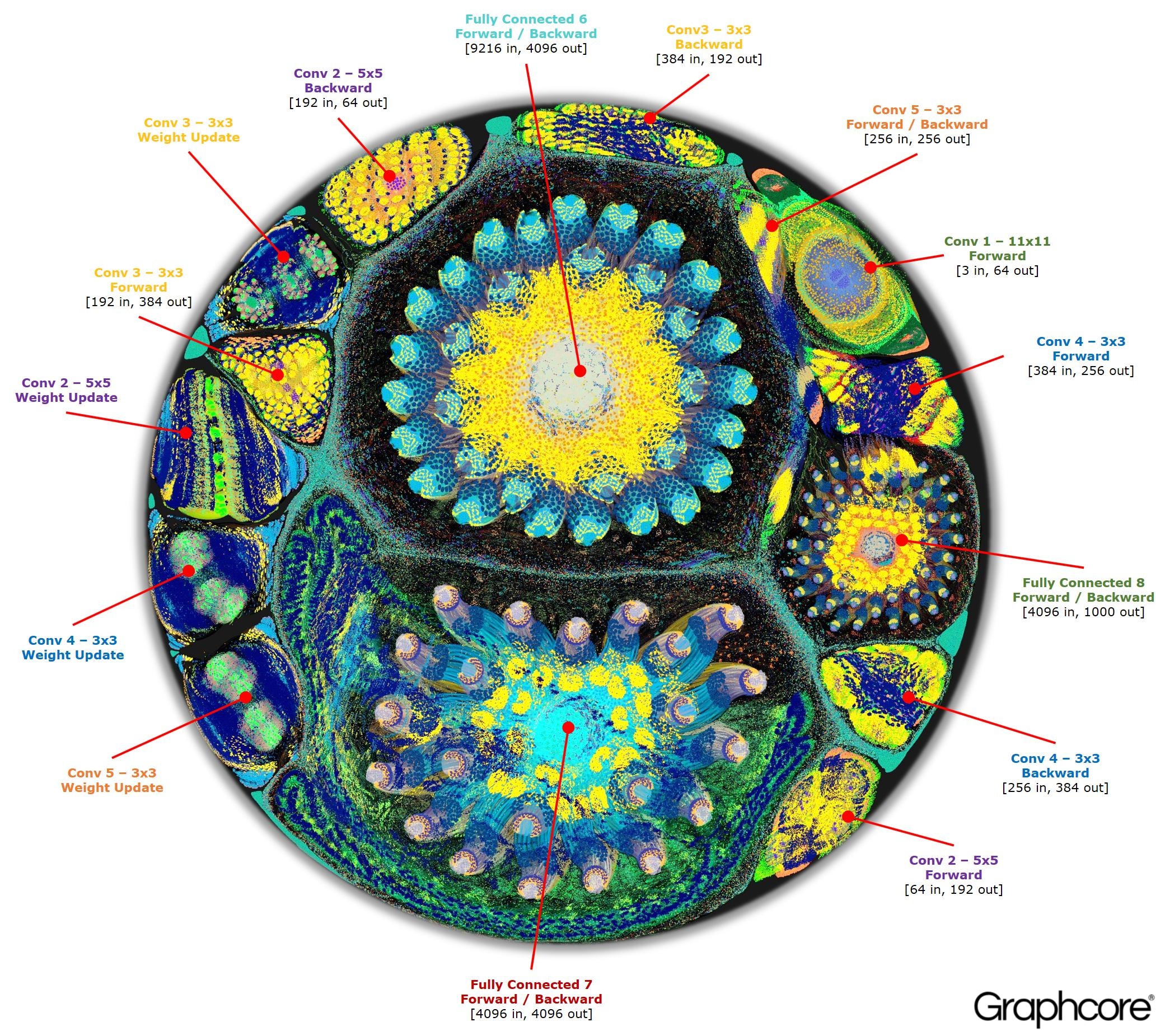}
\hfill
}
\caption{\label{graphcore_alexnet}
Graphcore's Poplar graph compiler turns the specification of an algorithm from a framework such as TensorFlow into a definition of the computation graph which is suitable for efficient distributed deployment on their IPU graph processor. This is a visualisation of the result for the AlexNet image classification CNN, supporting both training and run-time operation, where the spatial configuration and colouring indicates close connectivity of the different processing modules required.
  Image courtesy of Graphcore.
}
\vspace{2mm} \hrule
\end{figure}

\begin{figure}[t]
\centerline{
\hfill
\includegraphics[width=1.0\columnwidth]{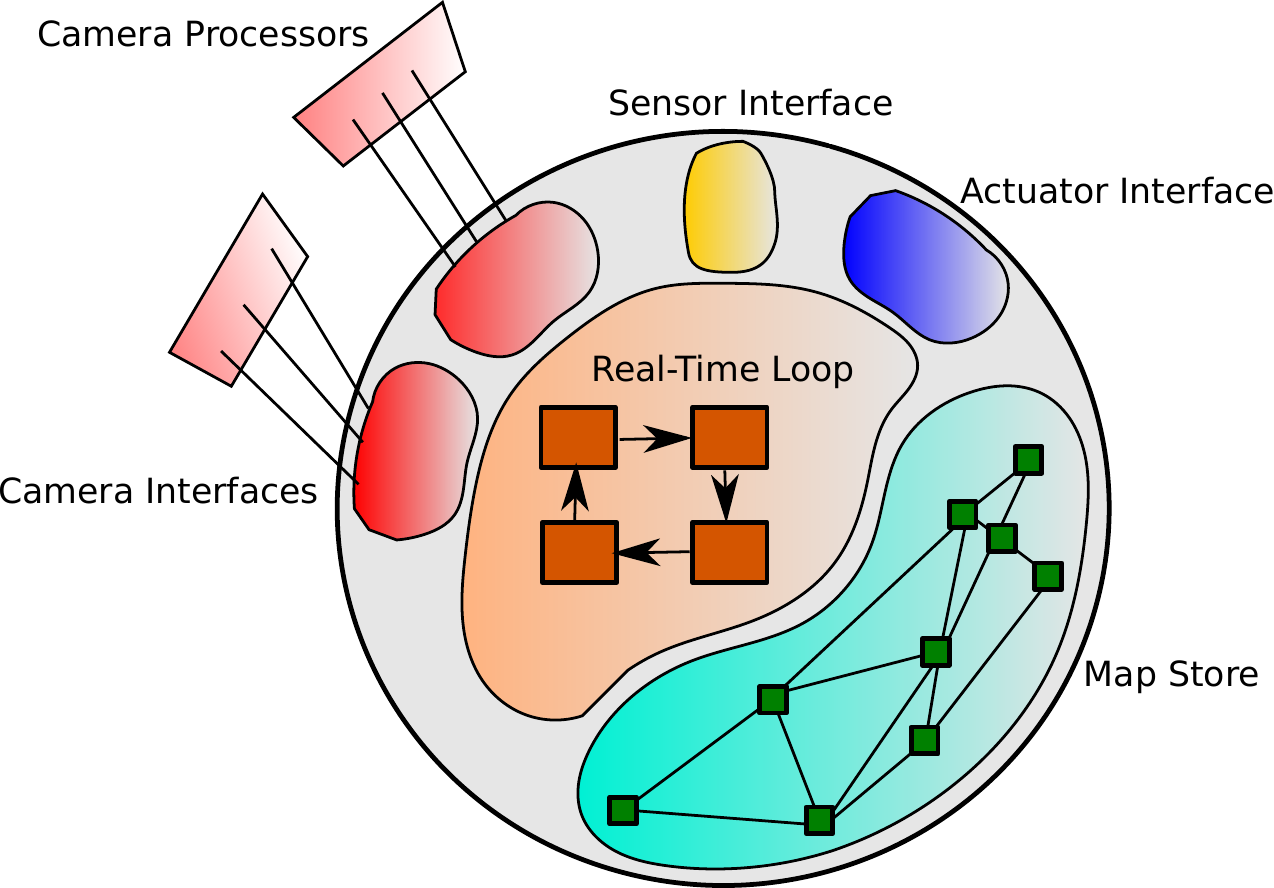}
\hfill
}
\caption{\label{fig:rvbrain}
Spatial AI brain: an imagining of how the representation and
processing graph structures of a general Spatial AI system might map
to a graph processor. The key elements we identify are the real-time
processing loop, the graph-based map store, and blocks which interface
with sensors and output actuators.
Note that we envision additional `close to the
sensor' processing built into visual sensors, aiming to reduce the data
bandwidth (eventually in two directions) between the main processor
and cameras, which will generally be located some distance away.
}
\vspace{2mm} \hrule
\end{figure}

We explained that in the overall architecture of our Spatial AI
computation system, the long term key to efficient performance is to match up
the natural graph structures of our algorithms
to the configuration of physical hardware. As we have seen, this
reasoning leads to the use of `close to the sensor' processing such as
in-plane image processing, and the attempt to minimise data transfer
from sensors and towards actuators or other outputs using principles
such as events.

However, we still believe that the bulk of
computation in an embedded Spatial AI system is best carried out by
a relatively centralised processing resource. The key reason for this
is the essential and ever-present role of an incrementally built and
used {\em world model} representing the system's knowlege of its state
and that of its environment. Every new piece of data from possibly
multiple cameras and sensors is ultimately
compared with this model, and either used to update it or discarded if
the data is not relevant to device's short or long-term goals.

What we are anticipating for this central processing resource is
however far from the model of a CPU and RAM-like memory store. A CPU
can access any contents of its RAM with a similar cost, but in Spatial
AI there is much more structure present, as we saw in
Section~\ref{section:graphs}, both in the locality of the data
representing knowledge of the world (\ref{section:geomgraphs}), and in the organisation of the
computation workload involved in incorporating new data (\ref{section:computegraphs}).

The graph structures of processing and storage should be built into
the design of the central world model computation unit, which should
combine storage and processing in a fully integrated way. There are
already significant efforts on new ways to design architectures which
are explicitly and flexibly graphlike.

To focus on one processor architecture currently under
development, Graphcore's IPU or graph processor is designed to efficiently carry out AI workloads which
are well modelled as operations on sparse graphs, and in particular
when all of the required storage is itself also held {\em within} the
same graph rather than in an external memory store. An IPU chip has a large
number of independent cores (in the thousands), each of which can
run its own arbitrary program and has its own local memory store, and
then a powerful communication substrate such that the cores can
efficiently send messages between themselves. When a program is to be
run on the IPU, it is first compiled into a suitable form by analysis
of the patterns of computation and communication it requires. A
suitable graph topology of optimised locations of operations, data,
and channels of communication is generated.

Graphcore have so far concentrated on how the IPU can be used to
implement deep neural networks, both at training time and
runtime. Figure~\ref{graphcore_alexnet} is a visualisation of the
graph structure identified by their graph compiler Poplar from the
TensorFlow definition of a standard CNN for image
classification. Starting from a neuron's elemental computations of
multiplying activations by weights, adding them together and passing
the result through a non-linearity, they build up a processing graph
which defines how these computations are joined together by data
inputs and outputs. When this very detailed graph is visualised at
increasing scale, with a view which `zooms out', clusters of highly
interlinked operations are apparent and these are concentrated and coloured in the
visualisation, and the whole graph is mapped to a disc. The result is
appealingly `brain-like', with the various layers of the network
shown as blocks of different sizes with their own internal structure visible.
It it important to note that this computational structure represents the operations of both training the CNN and using it for runtime operation.

Graphcore believe that the IPU will already have performance benefits
for executing well known deep neural networks compared to GPUs, and
their initial product offering will be an accelerator card for machine
learning in the cloud. However, their vision is greater than this, and
they say that the advantages of an IPU will become greater and greater
as the complexity of models increases. According to CEO Nigel Toon:
`the end game is deep, wide reinforcement learning, or more simply,
building networks that improve with use', thinking towards recurrent,
self-training structure with memory and constant input and output of data.
Clearly this is very well aligned with the vision we have for the
future of SLAM and Spatial AI, where we imagine systems with a
complex pattern of designed and learned modules, communicating in
real-time with input sensors and output actuators, and learning and
improving their internal representations continuously.

In Figure~\ref{fig:rvbrain} we have made a first attempt at drawing a
`Spatial AI brain' model which is somehow analagous to Graphcore's
visualisations. The disc contains the modules we anticipate running
inside the main processor. One of the two main areas is the map store, which
is where the current world model is stored. This has an internal graph
structure relating to the geometry of the world. It will also contain
significant internal processing capability to operate locally on the
data in the model, and we will discuss the role of this shortly.
The second main area is the real-time loop, which is where the main
real-time computation connecting the input image stream to the world
model is carried out. This has a processing graph structure and must
support large real-time data flows and parallel computation on image/map structures so is designed to optimise this.

The main processor also has additional modules. There are camera
interfaces, the job of each of which is
to model and predict the data arriving at the sensor to which it is
connected. This will then be connected to the camera itself, which the
physical design of a robot or other device may force to be relatively
far from the main processor. The connection may be serial or along
multiple parallel lines.

We then imagine that each camera will have its own `close-to-sensor' processing
capability built in, separated from the main processor by a data link.
The goal of modelling the input within the main processor is
to minimise actual data transfer to the close-to-sensor processors. It could be that the close-to-sensor processor performs purely image-driven computation, in a manner similar to the SCAMP project, and delivers an abstracted representation to the main processor. Or, there could be bi-directional
data transmission between the camera and main processor.
By sending model predictions to the close-to-sensor
processors, they know what is already available in the main
processor and should report only differences. This is a generalisation
of the event camera concept. An event camera reports only changes in
intensity, whereas a future optimally efficient camera should report
places where the received data is {\em different from what was predicted}.
We will discuss close-to-sensor processing further in Section~\ref{sec:close-to-sensor}.

\subsection{Map Store}

There is a large degree of choice possible in the representation of a
3D scene, but as explained in Section~\ref{sec:mapgraph}, we envision
maps which consist of graphs of learned features, which are linked in
multi-scale patterns relating to camera motion. These features must
represent geometry as well as appearance, such that they can be used
to render a dense predicted view of the scene from a novel
viewpoint. It may be that they do not need to represent full
photometric appearance, and that a somewhat abstracted view is
sufficient as long as it captures geometric detail.

Within the main processor, a major area will be devoted to storing this map, in a manner which is distributed around potentially a large number of individual cores which are strongly connected in a topology to mirror the map graph topology. In SLAM, of course the map is defined and grown dynamically, so the graph within the processor must either be able to change dynamically as well, or must be initially defined with a large unused capacity which is filled as SLAM progresses.

Importantly, a significant portion of the processing associated with
large scale SLAM can be built directly into this graph. This is mainly
the sort of `maintenance' processing via which the map optimises and
refines itself; including:
\bi
\item Feature clustering; object segmentation and identification.
\item Loop closure detection.
\item Loop closure optimisation.
\item Map regularisation (smoothing).
\item Unsupervised clustering to discover new semantic categories.
  \ei

With time, data and processing, a map which starts off as dense geometry and
low level features can be refined towards an efficient object level map.
Some of these operations will run with very high parallelism, as each
part of the map is refined on its local core(s), while other
operations such as loop closure detection and optimisation will
require message passing around large parts of the graph.
Still, importantly, they can take place in a manner which is internal to the map store itself.

The on-chip memory of next generation graph processors like
Graphcore's IPU is fully distributed among the processing tiles, and
the total capacity will initially not be huge (certainly when compared
with standard off-processor RAM), and therefore there should be an
emphasis on rapidly abstracting the map store towards an efficient
long-term form.

Optimising the geometric estimates in a SLAM map, such that the metric
map state is the most probably solution given the history of
measurements obtained, is a very well known optimisation problem known
as Bundle Adjustment (BA) in computer vision or graph optimisation in
robotics. The sparse constraints between poses and features due to
measurements lead to sparsity in the solvers needed
(e.g.~\cite{Konolige:BMVC2010}). While BA can be interpreted in terms
of matrix operations, it is also commonly posed directly as a graph
algorithm~\cite{Dellaert:IJRR2006}, and is clearly well suited to
implementation in a message passing manner on a graph processor.

\subsection{Real-Time Loop}

This other major part of the graph encompasses the core processing
which operates on live data input from cameras and other sensors and
connects it to the map. New images are tracked against projections
from the map and fused into updated representations. This is generally
massively parallel processing which is familiar from GPU-accelerated
dense SLAM systems, and these functions can be defined as fixed
elements in a computational graph which use a number of nodes and access a significant
fraction of the main computational resource. Functions such as
segmentation and labelling of input images will also be implemented
here (or possibly outside of the main processor by close-to-sensor
processing).

Data from the map store is needed for {\em rendering}, when a predicted view
of the scene is needed for tracking against new image data, and for
{\em fusion}, when information (geometry and labels) acquired from the
new data is used to update the map contents.

The most difficult issue in applying graph processing to the real-time
loop is the fact that the relevant part of the graph-based map store
for these operations changes continuously due to camera motion.  This
seems to preclude defining an efficient, fixed data path to the
distributed memory where map information will be stored.
Although there will be internal processing happening in the map store,
this will be focused on maintenance and it does not seem appropriate
to mode data rapidly around the map store, for instance such that the
currently observed part of the map is always available in a particular
graph location.

Instead, a possible
solution is to define special interface nodes which sit between the
real-time loop block and the map store. These are nodes focused on
communication, which are connected to the relevant components of
real-time loop processing and then also to various sites in the
map graph, and may have some analogue in the hippocampus of mammal brains.
If the map store is organised such that it has a `small
world' topology, meaning that any part is joined to any other part by
a small number of edge hops, then the interface nodes should be able
to access (copy) any relevant map data in a small number of
operations and serve them up to the real-time loop.

Each node in the map store will also have to play some part in this
communication procedure, where it will sometimes be used as part of
the route for copying map data backwards and forwards.

\section{Processing Close to the Image Plane}
\label{sec:close-to-sensor}

A robot or other device will have one or more cameras which interface
with the main processor, and we believe that the technology will
develop to allow a significant amount of processing to occur either
within the sensors themselves or nearby, with the key aim of reducing
the amount of redundant data which flows from the cameras.  A first
thought is to directly attach cameras to the main processor itself,
with direct parallel connections (wires) from the pixels to multiple
processing nodes, and certainly this is very appealing and could be
possible in some cases.  However, it is likely that in most devices
there are good reasons why the main processor will not be located
right next to the cameras, such as heat dissipation or space. The
ideal locations for cameras are unlikely to be the same as the ideal
location for a main processor. In any case, most future devices are
likely to have multiple cameras which all need to interface with the same
main processor and map store. Therefore some long camera to main
processor connections will be needed, and this motivates additional processing in the camera itself or very nearby.

Most straightforwardly, a sensor with in-plane processing similar to
SCAMP5~\cite{Martel:etal:ASRMOV2016} could be used to carry out purely
bottom-up processing of the input image stream; abstracting,
simplifying and detecting features to reduce it to a more compact,
data-rich form. Calculations such as local tracking (e.g. optical flow
estimation), segmentation and simple labelling could also be
performed.

Tracking using in-plane processing is an interesting problem. In plane
processing is good for problems where data access can be kept very
local, so estimating local image motion (optical flow), where the
output is a motion vector at every pixel, is well suited. At each
update, the amount of image change locally can be augmented using
local regularisation where smoothness is applied based on the
differences of neighbouring pixels.
If we look at the parallel implementation of an algorithm like Pock's
TV-L1 optical flow~\cite{Pock:PHD2008}, we see that it involves
pixelwise-parallel operations, where purely parallel steps relating to
the data term alternate with regularisation steps involving gradient
computation, which could be achieved using message passing between
adjacent neighbours. So such an algorithm is an excellent candidate
for implementation on an in-plane processor.

More challenging is the tracking usually required in SLAM,
where from local image changes we wish to estimate consistent global motion
parameters relating to a model. This could be instantaneous camera
motion estimation, where we wish to estimate the amount of global
rotation for instance between one frame and the next via whole image
alignment~\cite{Lucas:Kanade:IJCAI1981}, or tracking against a
persistent scene model as in dense
SLAM~\cite{Lovegrove:PHD2011,Newcombe:PHD2012}. When dense tracking is
implemented on standard processors, it involves alternation of purely
parallel steps for error term computation across all pixels with
reduction steps where all errors are summed and the global motion
parameters are updated. The reduction step, where a global model is
imposed, plays the role of regularisation, but the big difference is
that the regularisation here is global rather than local.

In a modern system, such global tracking is usually best
achieved by a combination of GPU and CPU, and therefore a regular (and expensive)
transfer of data between the two.
But we do not have this option if we wish to use in-plane processing
and keep all computations and memory local. Our main option for not
giving up on data locality is to give up on guaranteed global
consistency of our tracking solution, but to aim to converge towards
this via local message passing. Each pixel could keep its own estimate
of the global motion parameters of interest, and after each iteration
share these estimates with local neighbours. We would expect that
global convergence would eventually be reached, but that after a
certain number of iteration that the values held by each pixel could
be close enough that any single pixel could be queried for a usable
estimate. Convergence would be much faster if the in-plane processor
also featured some longer data-passing links between pixels, or more
generally had a `small world' pattern of interconnections.

Turning to the questions of labelling using local processing, this is certainly
feasible but a problem with sophisticated labelling is that current in
plane processor designs have very small amounts of memory, which makes
it difficult to store learned convolutional masks or similar.

Future processors for bottom-up processing may move beyond
operating purely in the image plane, and use a 3D stacking approach 
which could be well suited to implementing the layers of a
CNN. Currently 3D silicon stacks are hard to manufacture, and there are 
particular challenges around heat dissipation and cost, but the time
will surely come when extracting a feature hierarchy is a built-in
capability of a computer vision camera. Work such as
Czarnowski~\etal's {\bf featuremetric} tracking
~\cite{Czarnowski:etal:ICCVW2017} shows the wide general use this
would have. We should investigate the full range of outputs that a single
purely feed-forward CNN to do when trained with a multi-task learning approach.

An interesting question is whether processing close to the image plane
will remain purely bottom-up, or whether two-way communication
between cameras  and the main processor will be worthwhile. This would
enable model predictions from the world map to be delivered to the
camera at some rate, and therefore for higher level processing to be carried out
there such as model-based tracking of the camera's own motion or known objects.

In the limit, if a fully model-based prediction is able to be communicated to the camera, then the camera need only return information which is {\em different} from the prediction. This is in some sense a limit of the way that an event camera works. An event camera outputs data if a pixel changes in brightness --- which is like assuming that the default is that the camera's view of the scene will stay the same. A general `model-based event camera' will output data if something happens which differs from its prediction.
These questions come down to key issues of `bottom-up vs. top-down'
processing, and we will consider them further in the next section.

As a final note here,
any Spatial AI system must ultimately deliver a task-determined
output. This could be the commands sent to robot actuators,
communication to be sent to another robot or annotations and displays
to be sent to a human operator in an IA setting. Just as `close to the
sensor' processing is efficient, there should alse be a role for
`close to the actuator' processing, particularly because actuators or
communication channels have their own types of sensing (torque
feedback for actuators; perhaps eye tracking or other measures for an
AR display)  which need to be taken account of and fused in tight loops.

\section{Attention Mechanisms, or the Return of Active Vision}

The active vision paradigm~\cite{Ballard:AI91} advocates using sensing resources, such as
the directions that a camera points towards or the processing applied
to its feed, in a way which is controlled depending on the task at
hand and prior knowledge available. This `top-down' approach contrasts
with `bottom-up', blanket processing of all of the data received from
a camera.

In Davison's `Active Search for Real-Time Vision' the argument was
made that image processing in tracking and SLAM can be greatly reduced via prediction and
active search \cite{Davison:ICCV2005}. Image features need not be
detected `bottom up' across every frame, but their positions predicted based on a model
and searched for in a focused manner.  The problem with this idea was
that while image processing operations can certainly be saved by
prediction and active search, too much computation is required to
decide {\em where to look}. 
Most successful real-time vision systems since this time have instead used a `brute force' approach to low level image processing. In SLAM, this has meant either full-frame detection of simple features (e.g. FAST features~\cite{Rosten:Drummond:ECCV2006} as in PTAM~\cite{Klein:Murray:ISMAR2007}), or dense whole-image tracking as (e.g. DTAM~\cite{Newcombe:etal:ICCV2011}). 
The probabilistic calculations in~\cite{Davison:ICCV2005} were
sequential, and in particular difficult to transfer to the parallel
architectures taken advantage of by~\cite{Klein:Murray:ISMAR2007}
or~\cite{Newcombe:etal:ICCV2011}. 

Biological vision has a combination of bottom-up processing (early vision, which seems to operate in a purely bottom-up manner), and active, or top-down vision (proven by our moving heads and eyes which seek out relevant information for tasks based on model predictions).
We believe that an active vision approach will return to real-time
computer vision, but {\em at a higher level} than raw pixels. In
particular, the flexibility of graph processing architectures will be
much more amenable to active processing than CPU/GPU systems.

We understand very well now that entirely bottom-up processing of
an image or short sequence, via a CNN for example, can achieve remarkable things such as semantic labelling, object detection, depth estimation and local motion estimation.
At what level though should we interrupt this processing to bring in model-based predictions or task-dependent requirements?
For instance,  is it necessary to apply generic object detection processing to every image frame, when we can project previously mapped 3D objects into the newly estimated camera pose, and then apply  any detection effort only to image areas so far unexplained by good object models
(as shown by SLAM++~\cite{Salas-Moreno:etal:CVPR2013})?
The big question is: {\em What is the range of processing that we
  should expect a purely bottom-up algorithm to perform on input before it interacts with model-based data, predictions, and optimisation?}

The answer to this question must come back to the same kind of
information gain versus computation arguments which were used in
\cite{Davison:ICCV2005}, but now considered more broadly. That paper
studied low level feature matching in the context of a model-based
prediction (such as in incrememtal probabilistic SLAM), and showed that the amount of image processing required to match a set of
features could be much reduced by a one-by-one search strategy. Each search
reduced the total model uncertainty, and therefore also the size of
the search regions for the other features and the computation needed
to match them. The order of search required was determined based on
calculation of mutual information: the expected reduction in
uncertainty for each candidate measurement. The problem with this
method was that the reduction in image processing was ultimately
outweighed by the extra computation needed to make the probabilistic
and information theoretic calculations about what to measure next.  In
the end the elemental image processing computations needed to match
point features are not expensive enough to make active search
worthwhile, and it turns out to be better simply to detect and match
all features.

But as we reach towards full scene understanding from vision in real-time,
with the higher computational load required by elements such as dense
reconstruction, segmentation or recognition, active choices will make
sense again. Consider a simple possible bottom-up processing pipeline for semantic vision:
\be
\item Localisation using sparse features.
\item Dense reconstruction.
\item Pixel-wise scene labelling.
\item Object instance segmentation.
\item 3D object model fitting.
\ee
In analysis, we can estimate the per-pixel computational cost of each of
these elements as part of the pipeline, and assess the quality of
their output.
These measure should then be compared with the alternative `predict-update'
procedure, which could be carried out at any of the levels of output.
Here we recover existing models
from memory, warp them based on updates at the lower levels to make
predictions for the current time-step, and then use `fusion' processing on this
together with the new image to update and augment the predictions into
a final output. Computation should be focused on
filling in not yet modelled regions, 
such as new parts of the scene which come into
view at the edges of images or are revealed as occlusions are passed,
or improving the estimates of regions where estimates have high
uncertainty, such as those
seen
in more detail when approached or observed in improved lighting.
This `improvement' will usually take the form of a probabilistic
optimisation, where a weighted combination is made of model-based
predictions and information from the new data.

It is important that when assessing the relative efficiency of
bottom-up versus top-down vision, we take into account not just
processing operations but also data transfer, and to what extent
memory locality and graph-based computation can be achieved by each
alternative. This will make certain possibilities in model-based
prediction unattractive, such as searching large global maps or
databases.
The amazing advances in CNN-based vision means that we have to raise
our sights when it comes to what we can expect from pure bottom-up
image processing. But also, graph processing will surely
permit new ways to store and retrieve model data efficiently, and will
favour keeping and updating world models (such as graph SLAM maps)
which have data locality.






\section{Performance Metrics and Evaluation}

\label{section:metrics}

One of the hypotheses we made at the start of this paper is that the
usefulness of a Spatial AI system for a wide range of tasks is well
represented by a relatively small number of performance measures which
have general importance.

The most common focus in performance
measurement in SLAM is on localisation accuracy, and there have been
several efforts to create benchmark datasets for this
(e.g.~\cite{Sturm:etal:IROS2012}). An external pose measurement from a
motion capture system is considered as the ground truth against which a SLAM 
algorithm is compared.

We have argued that Spatial AI is about much more than pose
estimation, and more recent datasets have tried to broaden the scope
of what can be evaluated. Dense scene modelling is difficult to
evaluate because it requires an expensive and time-consuming
process such as detailed laser scanning to capture a complete model of
a real scene which is accurate and complete enough to be considered
ground truth. The alternative, for this and other axes of evaluation, is
to generate synthetic test data using computer graphics, such as the
ICL-NUIM dataset~\cite{Handa:etal:ICRA2014}.  Recently this approach
has been extended also to provide ground truth for semantic labelling
(e.g. SceneNet RGB-D~\cite{McCormac:etal:ICCV2017}), another output
where it is difficult to get good ground truth for real data.
It is natural to be suspicious of the value of evaluation against
synthetic test data, and there are many new approaches to gathering
large scale real mapping data, such as crowdsourcing
(e.g.~ScanNet~\cite{Dai:etal:CVPR2017}). However, synthetic data is
getting better all the time and we believe will only grow in importance.

Stepping back to a bigger point, we can question the value of using
benchmarks at all for Spatial AI systems. It is an often heard
comment that computer vision researchers are hooked on
dataset evaluation, and that far too much effort has been spent on
optimising and combining algorithms to achieve a few more percentage
points on benchmarks rather than working on new ideas and
techniques. SLAM, due to its real-time nature, and wide range of
useful outputs and performance levels for different applications, has
been particularly difficult to capture by meaningful benchmarks. 
I have usually felt that more can be learned about the usefulness of a visual
SLAM system by playing with it for a minute or so, adaptively and
qualitatively checking its behaviour via live visualisations, than from
its measured performance against any benchmark available. Progress has
therefore been more meaningfully represented by the progress of high
quality real-time open source SLAM reseach systems (e.g. MonoSLAM~\cite{Davison:ICCV2003,Davison:etal:PAMI2007}, PTAM~\cite{Klein:Murray:ISMAR2007},
KinectFusion~\cite{Newcombe:etal:ISMAR2011},
LSD-SLAM~\cite{Engel:etal:ECCV2014},
ORB-SLAM~\cite{Mur-Artal:etal:TRO2015}, OKVIS~\cite{Leutenegger:etal:IJRR2014}, SVO~\cite{Forster:etal:ICRA2014}, ElasticFusion~\cite{Whelan:etal:RSS2015}) that people can
experiment with, rather than benchmarks.

Benchmarks for SLAM have been unsatisfactory because they make
assumptions about the scene type and shape, camera and other sensor
choices and placement, frame-rate and resolution, etc., and focus in
on certain evalution aspects such as accuracy while downplaying other
arguably more important ones such as efficiency or robustness.
For instance, many papers evaluating algorithms against accuracy
benchmarks make choices among the test sequences available in a
dataset such as~\cite{Sturm:etal:IROS2012} and report performance only
on those where they basically `work'.

This brings us to ask whether we should build benchmarks and aim to
evaluate and compare SLAM systems at all? We still argue yes, but the
focus on broadening what is meant by a benchmark for a Spatial AI
system, and an acknowledgement that we should not put too much faith
in what they tell us.

The SLAMBench framework released by the PAMELA research
project~\cite{Nardi:etal:ICRA2015} represents important initial work
on looking at the performance of a whole Spatial AI system. In
SLAMBench, a SLAM algorithm (specifically
KinectFusion~\cite{Newcombe:etal:ISMAR2011}) is measured in terms of
both accuracy and computational cost across a range of processor
platforms and using different language implementations. SLAMBench2~\cite{Bodin:etal:ICRA2018}, recently released, now allows a a wide range of SLAM algorithms to be compared within a unified test framework, and other work in the PAMELA project is broadening this effort in other directions.
Saeedi~\etal~\cite{Saeedi:etal:ICRA2017}
have for instance started to address the issue of quantifying the SLAM
challenge represented by the type of motion and typical scene geometry
and appearance which is present in particular applications. The motion
of wearable devices, ground robots and drones are all very different,
and they are required to work in different environments. While SLAM on
a drone is certainly challenging due to rapid motion, a small ground robot can
also be a difficult case because of the low texture available in many
indoor environments and its position close to the ground where there
are many occlusions.

Over the longer term, we believe that benchmarking should move towards
measures which have the general ability to predict performance on
tasks that a Spatial AI might need to perform. This will clearly be
a multi-objective set of metrics, and analysis of Pareto
fronts~\cite{Nardi:etal:ARXIV2017} will permit choices to be made for a particular
application. 

A possible set of metrics includes:
\bi
\item Local pose accuracy in newly explored ares (visual odometry
  drift rate).
\item Long term metric pose repeatability in well mapped areas.
\item Tracking robustness percentage.
\item Relocalisation robustness percentage.
\item SLAM system latency.
\item Dense distance prediction accuracy at every pixel.
\item Object segmentation accuracy.
\item Object classification accuracy.
\item AR pixel registration accuracy.
\item Scene change detection accuracy.
\item Power usage.
\item Data movement (bits$\times$millimetres).
\ei

\section{Conclusions}

To conclude, the research area of Spatial AI and
SLAM will continue to gain in importance, and evolve towards the
general 3D perception capability needed for many different types of
application with the full fruition of the combination of estimation
and machine learning techniques. However, wide use in real
applications will require this advance in capability to be accompanied
by driving down the resources required, and this needs joined-up
thinking about algorithms, processors and sensors. The key to
efficient systems will be to identify and design for sparse graph
patterns in algorithms and data storage, and to fit this as closely as
possible to the new types of hardware which will soon gain in importance.

\section*{Acknowledgements}

This paper represents my personal opinions and musings, but has
benefitted greatly from many discussions and collaborations over
recent years. In particular I would like to thank all of my
collaborators on the PAMELA Project funded by EPSRC Grant
EP/K008730/1, especially Paul Kelly, Steve Furber, Mike O'Boyle, Mikel
Luj\'{a}n, Bj\"{o}rn Franke, Graham Riley, Zeeshan Zia, Sajad Saeedi,
Luigi Nardi,
Bruno Bodin, Andy Nisbet and John Mawer.

I am also grateful to my other colleagues at the Dyson Robotics Lab at Imperial College, the
Robot Vision Group, SLAMcore, Dyson, and elsewhere with whom I have discussed
many of these ideas; especially Stefan Leutenegger, Jacek Zienkiewicz, Owen Nicholson, Hanme Kim, Charles Collis,
Mike Aldred, Rob Deaves, 
Walterio Mayol-Cuevas, Pablo Alcantarilla,
Richard Newcombe, David Moloney, Simon Knowles, Julien Martel, Piotr Dudek and
Matthew Cook.


{\small
\bibliographystyle{ieee}
\bibliography{robotvision}
}

\end{document}

%% file: symbols.tex
\newcommand{\ignore}[1]{}

\newcommand{\ba}{\begin{array}}
\newcommand{\ea}{\end{array}}
\newcommand{\bc}{\begin{center}}
\newcommand{\ec}{\end{center}}
\newcommand{\be}{\begin{enumerate}}
\newcommand{\ee}{\end{enumerate}}
\newcommand{\bea}{\begin{eqnarray}}
\newcommand{\eea}{\end{eqnarray}}
\newcommand{\beas}{\begin{eqnarray*}}
\newcommand{\eeas}{\end{eqnarray*}}
\newcommand{\beq}{\begin{equation}}
\newcommand{\eeq}{\end{equation}}
\newcommand{\bfig}{\begin{figure}}
\newcommand{\efig}{\end{figure}}
\newcommand{\bi}{\begin{itemize}}
\newcommand{\ei}{\end{itemize}}
\newcommand{\bpic}{\begin{picture}}
\newcommand{\epic}{\end{picture}}
\newcommand{\btabular}{\begin{tabular}}
\newcommand{\etabular}{\end{tabular}}
\newcommand{\btable}{\begin{table}}
\newcommand{\etable}{\end{table}}

\newcommand{\es}{\vfill
                 \rule[-6mm]{170mm}{0.7mm} \\
                 \redw{{\tiny
		  \hfill S-\theslide}}
                 \end{slide}}

\providecommand{\etal}{{\em et~al.}}

\def \hbar {{\bar{h}}}

\makeatletter
\renewcommand*\env@matrix[1][*\c@MaxMatrixCols c]{%
  \hskip -\arraycolsep
  \let\@ifnextchar\new@ifnextchar
  \array{#1}}
\makeatother